\pdfoutput=1

\documentclass[10pt,twocolumn,letterpaper]{article}

\usepackage[pagenumbers]{cvpr}              
\usepackage{amsmath}
\usepackage{amssymb}

\usepackage[dvipsnames]{xcolor}

\makeatletter
\renewcommand{\paragraph}{%
  \@startsection{paragraph}{4}%
  {\z@}{0.4ex \@plus 1ex \@minus .2ex}{-1em}%
  {\normalfont\normalsize\bfseries}%
}
\makeatother
\usepackage[accsupp]{axessibility}

\usepackage{scalerel,stackengine}
\stackMath
\newcommand\reallywidehat[1]{%
\savestack{\tmpbox}{\stretchto{%
  \scaleto{%
    \scalerel*[\widthof{\ensuremath{#1}}]{\kern-.6pt\bigwedge\kern-.6pt}%
    {\rule[-\textheight/2]{1ex}{\textheight}}
  }{\textheight}%
}{0.5ex}}%
\stackon[1pt]{#1}{\tmpbox}%
}

\newcommand{\norm}[1]{\left\lVert#1\right\rVert}

\newcommand{\RR}{\mathbb{R}}
\newcommand{\SO}{\mathbb{SO}\!\left(3\right)}
\newcommand{\SE}{\mathbb{SE}\!\left(3\right)}
\newcommand{\SIM}{\mathbb{Sim}\!\left(3\right)}

\newcommand{\Ob}{\mathcal{O}}
\newcommand{\XX}{\mathbf{X}}

\definecolor{cvprblue}{rgb}{0.21,0.49,0.74}
\usepackage[pagebackref,breaklinks,colorlinks,citecolor=cvprblue, linkcolor=magenta]{hyperref}

\def\confName{CVPR}
\def\confYear{2026}

\title{4D Primitive-Mâché: Glueing Primitives for Persistent 4D Scene Reconstruction}

\author{Kirill Mazur, \quad Marwan Taher, \quad Andrew J. Davison \\
Dyson Robotics Lab, Imperial College London\\
{\tt\small \{k.mazur21, m.taher, a.davison\}@imperial.ac.uk}
}

\begin{document}


\twocolumn[{
    \renewcommand\twocolumn[1][]{#1}%
    \maketitle
     \includegraphics[width=\textwidth]{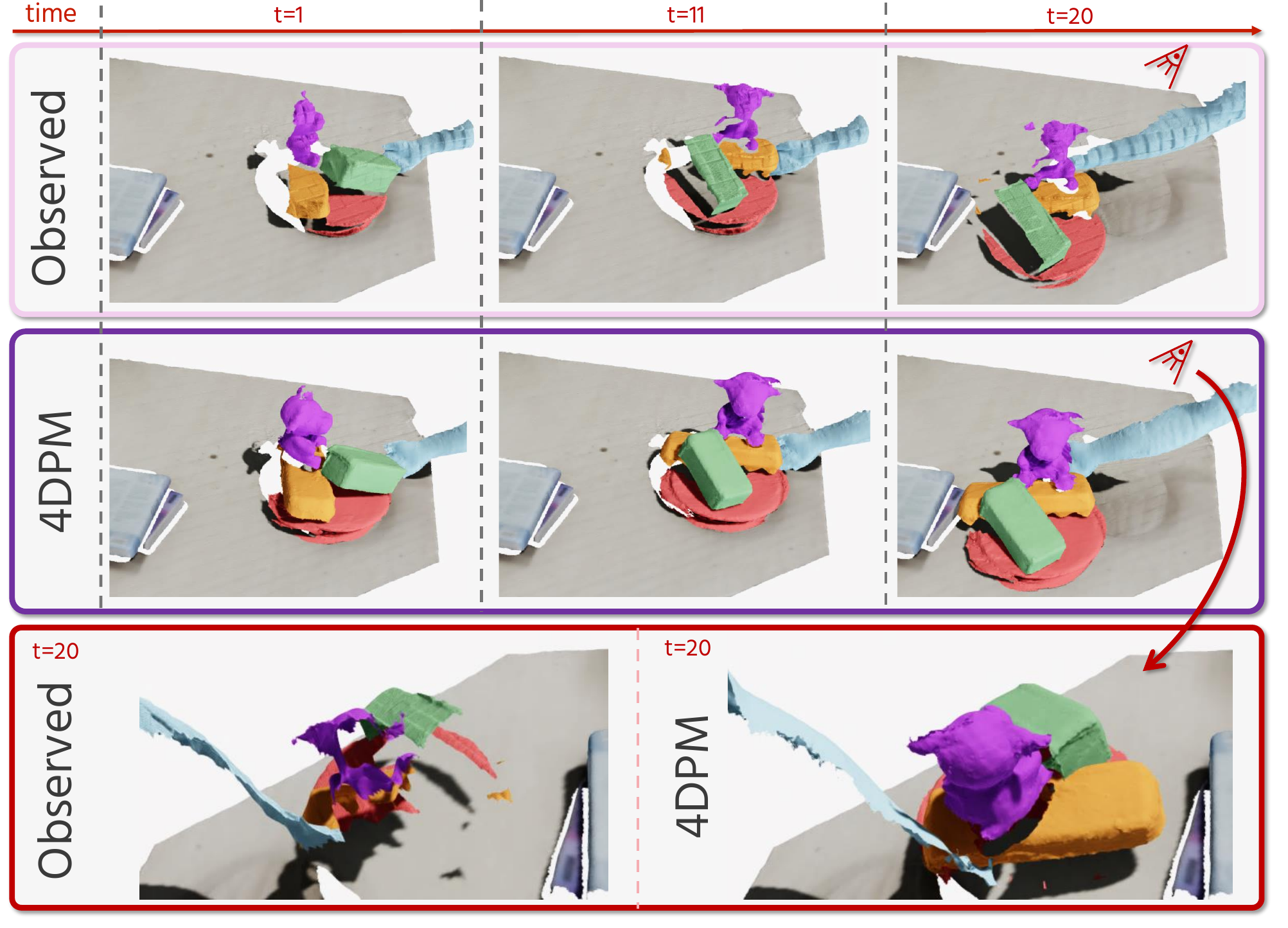}
  \captionof{figure}{Our method \textbf{(4DPM)} takes in casual monocular videos (captured by an iPhone) and outputs complete 3D scene reconstructions at \underline{every observed timestamp}, using all scene observations. The method takes in the outputs of a feedforward reconstruction model \textbf{(top row)} and glues dynamic geometry observations across time \textbf{(middle row)}. This results in a complete and accurate geometric reconstruction, which re-uses observations from all timestamps \textbf{(bottom row)}.}
  \label{fig:teaser}
  \vspace{2em}
}]

\begin{abstract}

We present a dynamic reconstruction system that receives a casual monocular RGB video as input, and outputs a complete and persistent reconstruction of the scene.  In other words, we reconstruct not only the the currently visible parts of the scene, but also all previously viewed parts, which enables replaying the complete reconstruction across all timesteps.

Our method decomposes the scene into a set of rigid 3D primitives, which are assumed to be moving throughout the scene. Using estimated dense 2D correspondences, we jointly infer the rigid motion of these primitives through an optimisation pipeline, yielding a 4D reconstruction of the scene, i.e. providing 3D geometry dynamically moving through time. To achieve this, we also introduce a mechanism to extrapolate motion for objects that become invisible, employing motion-grouping techniques to maintain continuity.

The resulting system enables 4D spatio-temporal awareness, offering capabilities such as replayable 3D reconstructions of articulated objects through time, multi-object scanning, and object permanence. On object scanning and multi-object datasets, our system significantly outperforms existing methods both quantitatively and qualitatively. 
Project page: \hyperlink{https://makezur.github.io/4DPM/}{https://makezur.github.io/4DPM/}

\end{abstract}

Building accurate and complete geometric reconstructions of dynamic scenes is a critical task in computer vision with broad applications in robotics, embodied intelligence, and augmented reality. Spatial AI systems typically operate in dynamic environments, requiring maps that are both complete and precise. 
Whilst SLAM~\cite{Davison:etal:PAMI2007, Klein:Murray:ISMAR2007, ORBSLAM3_TRO, murai2025mast3r} and SfM~\cite{schoenberger:etal:CVPR16, wang2025vggt} methods excel in building maps of static environments, they abstain from mapping dynamic scene parts. Agents typically interact with and modify their surroundings, and many environments are inherently dynamic, with objects and actors constantly changing position and configuration.

The main goal of mapping is persistence, i.e to retain as much relevant observed information as possible. For dynamic scenes, only the latest frame depth represents up-to-date geometric observations. Most dynamic mapping systems lack persistence and do not aggregate observed geometry across time. Thus, a lot of the information about previously observed scene elements is lost. Our goal in this paper is to build the most complete and persistent reconstruction possible, by not discarding previously observed information about moving objects.

The most general case of persistent, non-rigid dynamic reconstruction remains very challenging even for RGB-D systems and can only be performed on short sequences in controlled scenarios. Here, we make a piecewise-rigid motion assumption but show that this still covers a wide range of scenes and allows for replayable moving reconstructions consisting of many parts. 

We present a new scene reconstruction approach, 4D Primitive-Mâché (4DPM), which enables corresponding reconstruction of every observed keyframe \textit{at all given timestamps}. This contrasts with previous methods~\cite{zhang2024monst3r, li2025megasam, huang2025vipe} that reconstruct geometry (possibly coupled with point tracks) only at the time it was observed. This allows us to replay the 3D reconstruction in 4D.

The key to our method is to decompose scenes into object-like chunks, as in SuperPrimitive~\cite{Mazur:etal:CVPR2024} (SP), a primitive-based approach to SLAM and SfM. We tackle dynamic scene reconstruction using scene primitives as the underlying representation. This enables compact and optimisable representation of dense scene geometry. Whilst SP assumes pixel-aligned primitives and estimates only unknown depth scales, we extend this framework by endowing each primitive with rigid motion parameters.

This formulation allows us to efficiently reduce the complex all-to-all temporal dense mapping problem to estimating \textit{a single} $\SE$ \textit{pose per primitive}. The pose encodes the transformation to the last observed timestamp of that object. We demonstrate that this compact representation suffices to encode the complete temporal trajectory of each primitive, enabling tractable motion inference across a diverse range of dynamic scenes.

In order to estimate the motion of these primitives (``glue'' them across time), we employ an off-the-shelf dense 2D correspondence estimation network~\cite{harley2025alltracker}. Given the estimated correspondence between 3D primitives, we estimate their motion by direct 3D alignment in their respective coordinate systems. 

To briefly summarise our contributions:
\begin{itemize}
\item We introduce a primitive-based motion parameterisation that represents per-pixel motion fields through sparse per-primitive $\SE$ poses, dramatically reducing the dimensionality of dynamic scene reconstruction whilst maintaining expressiveness for piecewise-rigid motion.
\item We demonstrate geometric accuracy significantly higher than the existing monocular methods on challenging scenarios including multi-object interactions, validated against multi-view ground truth.
\item Our primitive persistence enables spatial memory capabilities, maintaining representations of temporarily occluded objects. To the best of the authors' knowledge, our method is the first monocular reconstruction system demonstrating such capabilities.
\end{itemize}

\section{Related Work}

\paragraph{Compact Scene Representations}
Rather than optimising dense geometric (such as depth) values for every observed pixel, compact representations parameterise scenes through learned models or structured primitives. 
Code-based methods~\cite{Bloesch:etal:CVPR2018, Czarnowski:etal:RAL2020, Matsuki:etal:RAL2021} learn depth prediction networks conditioned on optimisable latent codes, enabling joint optimisation during reconstruction. Similarly, COMO~\cite{dexheimer2024compact} employs explicit 3D control points as an optimisable scene representation. Primitive-based approaches take this further by representing scenes as compositions of higher-level structures: SLAM++~\cite{Salas-Moreno:etal:CVPR2013} represents scenes as collections of CAD models and optimises their poses, while SuperPrimitive~\cite{Mazur:etal:CVPR2024} decomposes frames into 2.5D primitives obtained through surface normal integration within image segments, estimating their depth scales through optimisation.

\paragraph{Compact Motion Representations}

Early sparse reconstruction systems of moving bodies segmented motion into multiple rigid components~\cite{costeira1995multi}, which was later extended to articulated objects~\cite{tresadern2005articulated}. To move beyond rigid-body assumptions, subsequent work explored compact motion representations through decomposition: \cite{Bregler2000RecoveringN3} proposed decomposing moving objects as linear combinations of shape bases, whilst \cite{akhter2008nonrigid} instead decomposed trajectories using a linear basis set. 

With the advent of commercially available depth sensors, fusion-based methods enabled dense dynamic reconstruction. DynamicFusion~\cite{Newcombe:etal:CVPR2015} pioneered RGB-D non-rigid reconstruction by representing motion via a sparse set of deformation nodes and fusing observations into a canonical, voxel-based TSDF model. 
Object-centric approaches such as Co-fusion~\cite{Runz::Agapito::ICRA2017} and MID-fusion~\cite{Xu:etal:ICRA2019} spawned individual dense maps --- surfel-based and octree-based respectively --- for different objects. In contrast with~\cite{Newcombe:etal:CVPR2015}, these methods enabled dynamic scene level reconstruction. Whilst these methods share our object-based motion representation, they require continuous depth streams for map fusion and model-based tracking, making them inapplicable to monocular scenarios.

For monocular dense reconstruction, SuperPixel Soup~\cite{kumar2017monocular, kumar2019superpixel} represents scene motion by decomposing the scene into rigidly moving superpixels; however, these methods are largely limited to two-view, small-baseline configurations.

\paragraph{4D reconstruction methods}
In recent years, joint depth and camera pose estimation from casual videos containing dynamic objects has gained significant attention. Initial approaches combined learned priors with scene-specific optimisation~\cite{kopf2021robust, zhang2022structure, huang2025vipe, li2025megasam}, which were later followed by feed-forward reconstruction models~\cite{wang2025pi3, zhou2025page}. Our method is complementary to these approaches: whilst they estimate geometry only at the time of observation, we focus on remapping observed reconstructions to all  timestamps. Consequently, the outputs of these methods can be used as the input to ours.

Inspired by the recent success of DUSt3R~\cite{wang2024dust3r}, a series of recent methods~\cite{st4rtrack2025, sucar2025dynamic, Zhang_2025_ICCV} has extended DUSt3R's shared coordinate frame mapping paradigm to dynamic scenes by establishing temporal correspondences, thereby warping point maps across different time frames. While these methods are theoretically more expressive, real data for training supervision remains extremely sparse, leading to limited performance even on scenes with predominantly rigid motion. Additionally, pair-wise methods suffer from quadratic complexity when the number of frames increases.

\section{Method}

\begin{figure*}[h!]
    \centering
    \includegraphics[width=\linewidth]{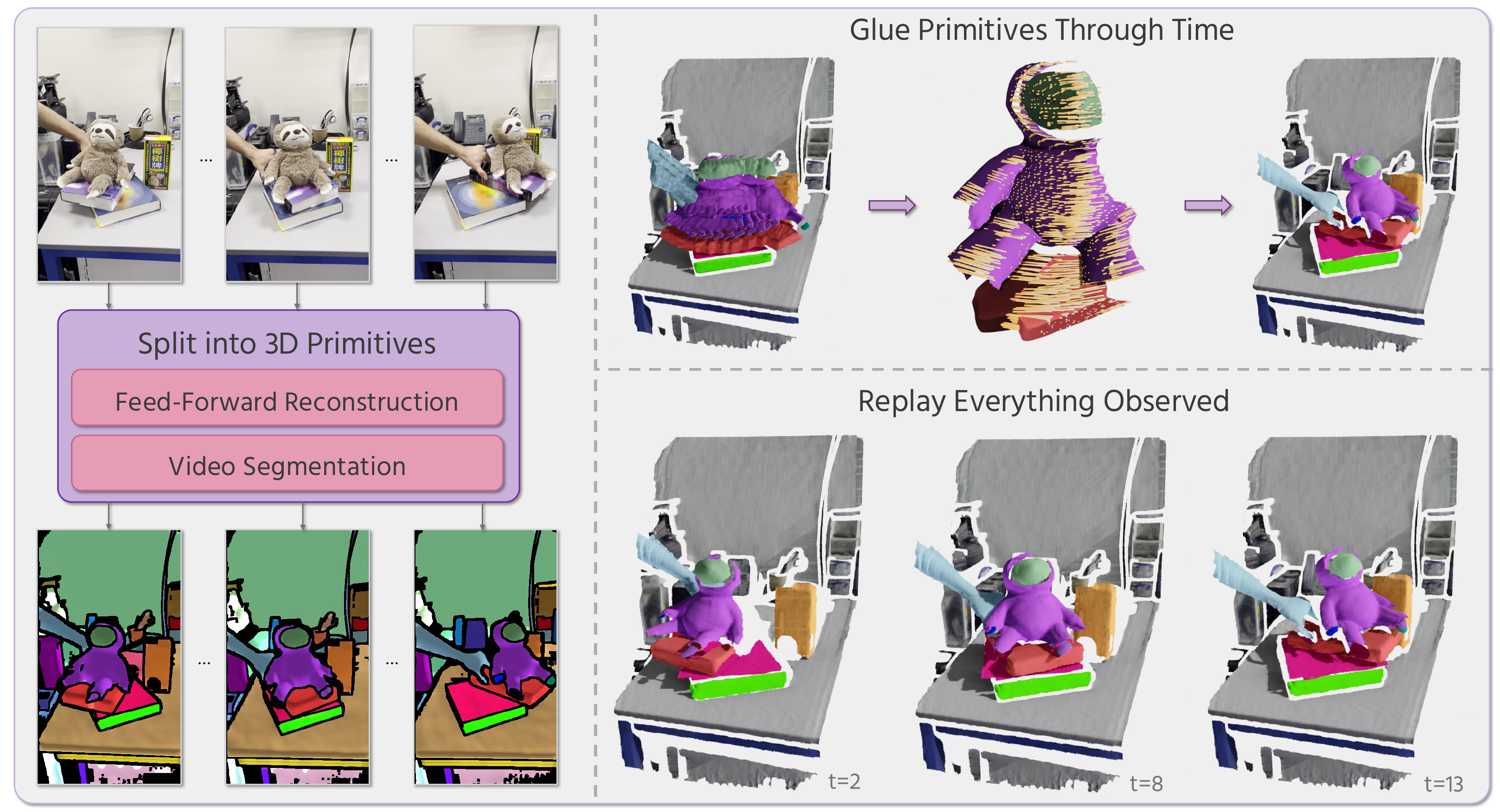}
    \caption{\textbf{4D reconstruction with 4DPM.} \textbf{(left)} Our frontend takes in a monocular RGB video and splits it into a set of 3D primitives. Each primitive is represented as a 3D point map in the world coordinate space, cut out by a segmentation mask. These primitives are matched across time (visualised with consistent colours) to form consistent entities across time, to which we refer as \textit{objects}. 
    \textbf{(top right)} Given geometric observations positioned at their respective timestamps, we ``glue'' primitives belonging to the same object across time according to their estimated dense 2D correspondences. 
    \textbf{(bottom right)} The resulting complete reconstruction can be replayed across all observed timestamps.}
    \vspace{-12pt}
    \label{fig:method}
\end{figure*}

\subsection{Formulation}
Our geometry representation is based on point maps $\XX_k^{t} \in \RR^{H \times W \times 3}$, which represent 3D geometry captured from camera $k$ and warped to time $t$ in world coordinates. For a video sequence $\mathcal{I} =  \{I_0, \dots, I_n\ | I_i \in \RR^{H \times W \times 3} \}$, the scene reconstruction at time $t$ is the collection $\mathcal{X}^t = \{\XX_0^t, \ldots, \XX_n^{t}\}$. 

Prior work reconstructs image-aligned point maps only at their respective observation times, yielding $\{\XX_0^0, \XX_1^1, \ldots, \XX_n^{n}\}$. 
In contrast, our goal is to reconstruct the complete scene geometry at every observed timestamp, obtaining $\{ \mathcal{X}^{0}, \ldots, \mathcal{X}^{n} \}$. We achieve this by taking observation-time point maps from a feed-forward model~\cite{wang2025pi3} as input to our 3D primitive glueing method.

\paragraph{Primitives} Every keyframe $I_i$ is partitioned into a set of non-overlapping image regions $S_p \in I_i$. These regions cut 3D primitives out of $\XX_i$, which we denote as $S_p \odot \XX_i$. To represent scene motion, we allow these primitives to move as rigid bodies over time, which is parametrised with $\SE$ poses $T({S_p})$ of each individual 3D primitive. In~\cref{sec:backend} we describe how to infer this motion from estimated 2D correspondences between images.

\paragraph{Pose representation} Object and camera poses $T \in \SE$ are stored as $4 \times 4$ matrices. For optimisation, we employ a Lie group parametrisation following the notation in~\cite{sola2018micro}. Pose updates are represented as Lie algebra elements $\tau \in \mathfrak{se}(3) \simeq \RR^6$, and transformations are updated via: 
\begin{equation}
    T \leftarrow  T \oplus \tau
\end{equation}

\subsection{Objects}
\label{sec:objects}
Our 3D primitives are matched across time to form object-like structures for two key reasons. First, this enables filtering of spurious correspondences, which frequently arise at object boundaries. Second, and more importantly, this matching facilitates a compact motion representation: each 3D primitive requires only a single $\SE$ to represent its motion at all visible timestamps, see ~\cref{sec:timeremap}.

The scene primitives are clustered into \textit{object} groups $\Ob $. Intuitively, our method assumes each object-like entity to be rigidly moving in the scene. Each object $\Ob$ is a set of primitives, grouped by time $\Ob = \{ S^{t_{start}}, \dots , S^{t_{end}} \}$.

To represent an object $\mathcal{O}$'s motion, we couple each segment $S \in \mathcal{O}$ with an $\SE$ pose $T(S)$, which maps it in the coordinate frame of the last observed segment, $S^{t_{end}}$. The pose $T({S^{t_{end}}})$ corresponding to the last observed segment is set to identity, since pose estimation for each object entity has a natural $\SE$ gauge freedom~\cite{Triggs:etal:VISALG1999, Strasdat:etal:RSS2010} (since the scale between objects is fixed).

\paragraph{Static-Dynamic classification} 

Monocular videos present a natural geometric ambiguity: from 2D correspondences alone, motion induced by camera movement cannot be easily disambiguated from actual scene motion. To resolve this gauge freedom in the backend optimisation, we freeze objects based on their initial correspondence residuals. These frozen objects are assumed to be static in the world frame, such that their observed motion arises solely from camera movement.

\begin{figure}[h]
    \centering
    \includegraphics[width=\linewidth]{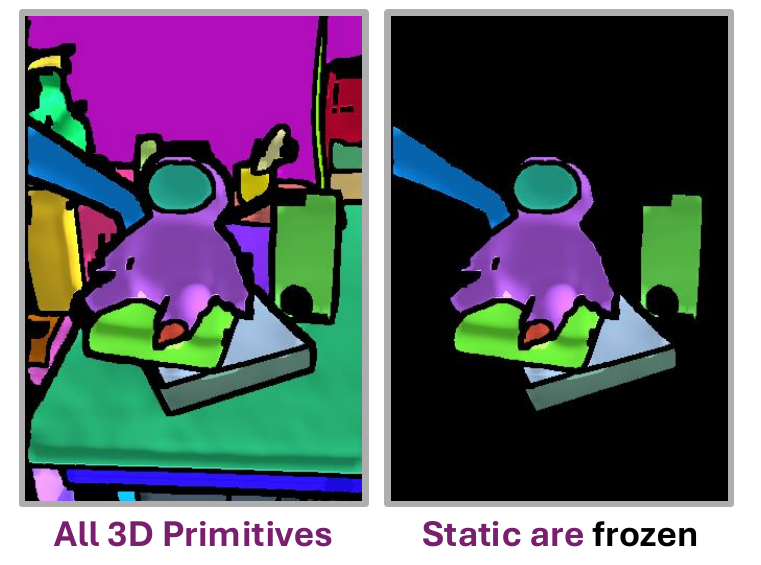}
    \caption{\textbf{Static vs dynamic segmentation.} We visualise all estimated primitives on the \textbf{left}. Before motion estimation, we freeze primitives with insufficiently high correspondence residuals, assuming they are static. On the \textbf{right}, only dynamic primitives are shown. Our system produces motion segmentation as a by-product.}
    \vspace{-16pt}
    \label{fig:static}
\end{figure}

\subsection{Frontend}
\paragraph{Geometry}
Given a set of keyframes $\mathcal{I}$, we run a feed-forward scene reconstruction model $\pi^3$~\cite{wang2025pi3}, which estimates point map at observation times  $\pi^3(\mathcal{I}) = \{\XX_0^0, \XX_1^1, \ldots, \XX_n^{n}\}$. This serves as the basis for our Primitive-Mâché glueing algorithm.

\paragraph{Segmentation} Given a video sequence $\mathcal{I}$, our frontend should decompose this video into a set of non-overlapping objects, covering most of the video sequence. 

To this end, we segment the first frame of the video as in SuperPrimitive~\cite{Mazur:etal:CVPR2024}. These masks are propagated to all other keyframes using SAMv2~\cite{ravi2024sam2}. Our algorithm sequentially processes subsequent frames, instantiating new objects in uncovered image regions via active sampling and propagating them forwards, akin to~\cite{zheng2025one}.

This process partitions the imagery into the set of objects introduced in~~\cref{sec:objects}. 

\paragraph{Correspondence}
To extract dense correspondence between subsequent keyframes, we run a dense point tracking model~\cite{harley2025alltracker} on the videos between the keyframes. These correspondences are then filtered to only align points that belong to the same object. 

\subsection{Backend}
\label{sec:backend}

To estimate 3D primitives' poses, we employ dense 2D correspondences between consecutive keyframes as constraints in a 3D optimisation. We perform joint optimisation across all objects and keyframes to find poses that best align corresponding 3D points. Intuitively, this process ``glues'' the objects out of their constituent primitives.

We use a dense correspondence network to estimate pixel-wise flow between temporally adjacent keyframes $I_k$ and $I_{k + 1}$. While non-adjacent keyframes can also be used, we find adjacent pairs provide the best efficiency-robustness trade-off. The network\cite{wang2025pi3} also provides per-pixel correspondence confidence weights which we use in the cost calculation.

Given a pointmap $\XX_k^k$ (which we further denote as $\XX_k$ for brevity)  and flow field, we obtain corresponding 3D points $\reallywidehat{\XX}_{k + 1}$ by warping $\XX_{k + 1}$ according to the flow. Similarly, segment masks $S_{k}$ are warped to produce $\reallywidehat{S_{k+1}}$.

For a single object $\Ob = \{S_n, ... , S_m \}$ , our dense optimisation aims to estimate the poses $T(S_n), \ldots , T(S_m)$ of its constituent primitives.  Our cost function directly minimises the distance between the 3D points of primitives belonging to the same object, weighted by correspondence confidence $\mathbf{w}_{ij}$:

\begin{equation}
E \left( \Ob \right) =  \sum_{(i, j) \in \mathcal{T}(\mathcal{\Ob}) } \norm{ \mathbf{w}_{ij} \cdot S_i \cdot \reallywidehat{S_j} \left( T_j^{-1} T_i \XX_i - \reallywidehat{\XX_j} \right)}_{\rho}
\end{equation}

where $\mathcal{T}(\mathcal{\Ob})$ is the set of temporally-adjacent primitives which belong to object $\Ob$ and $|| \cdot ||_{\rho}$ is the Huber cost function.

The final cost is combined across all objects present in the scene that are not classified as static:

\begin{equation}
E_{\text{final}} = \sum_i E(\Ob_i) 
\end{equation}

Jacobians $\mathbf{J}$ for all poses are derived analytically and efficiently inferred in parallel for all objects in the scene. We solve this optimisation problem for all objects jointly via iteratively reweighed least squares using Gauss-Newton optimisation~\cite{Triggs:etal:VISALG1999}: 

\begin{equation}
    \mathbf{J}^T \mathbf{W} \mathbf{J} \tau = - \mathbf{J}^T \mathbf{W} \mathbf{r}, \, \,  T_i = T_i \oplus \tau 
\end{equation}

\subsection{Time Remapping}
\label{sec:timeremap}
Having estimated poses for all primitives, we now aim to produce reconstructions of every frame at each timestamp, namely $\mathcal{X}^{0}, \ldots \mathcal{X}^{n}$. In contrast to the non-rigid scenario, having mapped rigid objects onto their latest observed frame allows us to infer these object' positions at all observed timestamps. 

More precisely, given an object $\Ob$ represented with primitives $\{ S^{t_{start}}, \dots , S^{t_{end}} \}$ together with their estimated poses $T(S^{t_{start}}), \dots T(S^{t_{end}}))$ we can express the warping as follows. As covered in~\cref{sec:objects}, each pose maps a primitive onto the coordinate system of the last observed primitive, $S^{t_{end}}$. Then, the warping of $S^{p}$ to time $q$ can be naturally expressed as: 
\begin{equation}
    T^{p \mapsto q}  \colon =  \left[ T(S^{q} ) \right] ^{-1} T(S^{p}\textbf{)}   ~,
\end{equation}
which corresponds to first transforming $S^{p}$ onto the coordinate frame of $S_{t_{end}}$ and then pulling it back to time $q$.

\subsection{Motion Segmentation}
\label{sec:motionseg}
While some objects might become unobservable, humans can still reason about their positions in the 3D space. Our representation enables such reasoning.

When an object becomes unobservable (e.g., an item placed in a closing drawer), we infer its continued motion (when possible) by linking it to a parent object that remains visible.

We identify potential parent objects using two criteria: spatial contact and velocity similarity. Parents are only assigned across objects in spatial contact, up to transitive closure with the preference to objects with clustered velocities.

\paragraph{Spatial Contact} 
To determine whether or not two objects are in contact, we fit Oriented Bounding Boxes (OBBs) to every object at each timestamp. We consider the objects to be \textit{in contact} if their bounding boxes extended by $\alpha$ have non-zero intersection. In our experiments, we set $\alpha = 1.1$. 
\paragraph{Velocity Clustering} 
Every object has its own coordinate system for poses, as discussed in~\cref{sec:objects}. Hence object poses $T_\text{obj}' = \mathcal{F} T_{obj}$ are defined up to an unknown gauge freedom $\mathcal{F} \in \SE$. Thus comparing their poses directly is ill-posed. However, comparing velocities is gauge-invariant:
\begin{equation}
\begin{split}
    T'(t)^{-1} T'(t - 1) &= T(t)^{-1} \mathcal{F}^{-1} \mathcal{F} T(t - 1) \\
    &= T(t)^{-1} T(t - 1)
\end{split}
\end{equation}
Velocities $\mathcal{V}$ and $\mathcal{W}$ of two co-observed objects are then compared  $\operatorname{log} (\mathcal{V}^{-1} \mathcal{W})$ under the Mahalanobis distance with diagonal covariance parameters $\sigma_{\tau}$ (translation) and $\sigma_{\psi}$ (rotation). Objects with distance below threshold are considered to have similar motion.

\section{Experiments}

We evaluate our dynamic geometry reconstruction quality, both in terms of accuracy and completeness. This is done by warping all observations to the final timestamp and comparing against the GT geometry of the scene. In this setup, we assume a scene is captured by a set synchronised RGB-D cameras, providing multi-view GT. Note that only one RGB camera stream is used as input to all methods, to which we further refer as the main camera.

Ideally, after processing the whole sequence, a method should reuse all observations to produce as complete and accurate reconstruction as possible. Specifically, for all methods, we time-warp each pointmap onto the latest keyframe~$n$, yielding the final reconstruction $ \mathcal{X}^n = \{\XX_0^n, \XX_1^n, \ldots, \XX_n^{n}\}$.

Due to gauge freedom, the estimated geometry has a $\SIM$ transformation ambiguity. For every method, we align its prediction to the ground truth coordinate system using Umeyama alignment~\cite{Umeyama1991LeastSquaresEO}, applying the transformation computed from the last observed keyframe's pointmap.

For DUSt3R-based baselines (St4track~\cite{st4rtrack2025} and POMATO~\cite{Zhang_2025_ICCV}), we first extract all pointmaps temporally aligned to the final observation $X_n^{n}$, and align them to a shared coordinate frame. 

\paragraph{Metrics} We evaluate reconstruction quality using two complementary metrics against pseudo-ground-truth pointmaps: (1) \textit{accuracy} --- the percentage of predicted points within a 1cm threshold of the ground truth, measuring precision; and (2) \textit{recall} --- the percentage of ground truth points covered by at least one predicted point within 1cm, measuring completeness. Following the established practice~\cite{Knapitsch2017}, we report the F-score, the harmonic mean of accuracy and recall, which reflects both metrics in a single value.

In all our experiments, we evaluate reconstruction quality solely on dynamic scene parts. Including static parts in the evaluation would dominate the completeness (i.e., recall) metric. Specifically, auxiliary cameras in multi-view datasets typically capture static scene regions that are never visible from the main camera. Evaluating completeness on these regions would penalise a method for failing to reconstruct geometry that was never observed in the input video, making the recall measure uninformative for assessing the quality of dynamic scene reconstruction.

Long sequences are split onto chunks of $150$~consecutive frames and all methods are tested on each chunk separately. In this case, we report average result per sequence. 

\paragraph{Baselines} 
We propose two natural baselines for our method. First, ($\pi^3$ last view in the tables) only uses the latest frame pointmap as estimated geometry. In the last frame the positioning of dynamic geometry is up to date, but it lacks completeness.   

Another natural baseline ($\pi^3$ in the tables) is simply the untouched estimate of a feed-forward reconstruction model run on the input keyframes. While the geometrical information in this case is complete, it is not properly positioned for early keyframes.

Besides these baselines, we also compare our system to state-of-the art methods, that are capable of dense time-warping of observed scene geometry: St4track~\cite{st4rtrack2025}, POMATO~\cite{Zhang_2025_ICCV} and TraceAnything~\cite{liu2025trace}.

All our experiments are conducted with a single NVIDIA GeForce RTX 4090 GPU.
\subsection{Object Scanning}

We evaluate our system on HO3D~\cite{hampali2020honnotate}, an object scanning dataset in which stationary cameras capture humans manipulating objects of interest. We select sequences recorded with four calibrated depth cameras, which provide ground truth depth for evaluation. Although human hands are non-rigid, hand poses remain largely static in this dataset. We therefore report geometric accuracy for both hands and scanned objects (i.e., dynamic scene parts) using the ground truth segmentation masks provided.

In~\cref{tab:ho3d}, we report per-sequence F-scores for all methods, alongside average F-score, recall, and precision. While ($\pi^3$ last frame) achieves higher precision, its recall (i.e., scene coverage) is poor because information from all keyframes except the last is discarded. Other baselines perform well in terms of recall, indicating decent object coverage. However, most fall short in precision. Our method provides the best balance between precision and recall, achieving the highest F-score across all compared methods.

\begin{table*}[h]
\begin{tabular}{l|ccccccc|c||cc}
\hline
                     & ABF1   & BB1    & GPMF1  & GSF1   & MDF1   & ShSu1  & SiBF1  & avg. F-score  & Precision                              &   Recall \\ 
\hline
$\pi^3$ last view    & 0.3185 & 0.3399 & 0.2553 & 0.4330 & 0.2604 & 0.2794 & 0.3581 & 0.3206   &     \textbf{0.9255}                         &   0.2018 \\
$\pi^3$              & 0.4405 & 0.4913 & 0.5257 & 0.5394 & 0.4855 & 0.6744 & 0.4968 & 0.5219   &      0.4735                                 &   0.6296 \\
\hline
St4Track             & 0.4717 & 0.3015 & 0.5699 & 0.6414 & 0.4913 & \textbf{0.7892} & 0.5095 & 0.5392   &        0.4549                               &   0.7293  \\
POMATO               & 0.4572 & 0.4304 & 0.5254 & 0.6747 & 0.5581 & 0.5271 & 0.4766 & 0.5214   &        0.4065                               &  \underline{0.7650}  \\
TraceAny             & \underline{0.5656} & \underline{0.6323} & \underline{0.6199} & \underline{0.7158} & \underline{0.5772} & 0.7375 & \underline{0.6069} & \underline{0.6365}   &        0.5748                               &   0.7398  \\ 
\hline
Ours           & \textbf{0.7259} & \textbf{0.7306} & \textbf{0.7645} & \textbf{0.7636} & \textbf{0.7586} & \underline{0.7758} & \textbf{0.7824} & \textbf{0.7573}   &  \underline{0.7630}               &    \textbf{0.7774} \\
\hline
\end{tabular}
\label{tab:ho3d}
\caption{\textbf{Quantitative evaluation on HO3D dataset.} We report F-score (threshold at 1 cm) per sequence for all methods. Average F-score, precision, and recall across all sequences are also reported. Our method marginally outperforms all baselines in terms of F-score for dynamic object scanning, providing the best balance between completeness and accuracy. Best is highlighted as \textbf{bold}, while second-best is \underline{underscored}.}
\end{table*}

\subsection{Multi-Object Dataset}
\begin{figure*}[h]
    \centering
    \includegraphics[width=\linewidth]{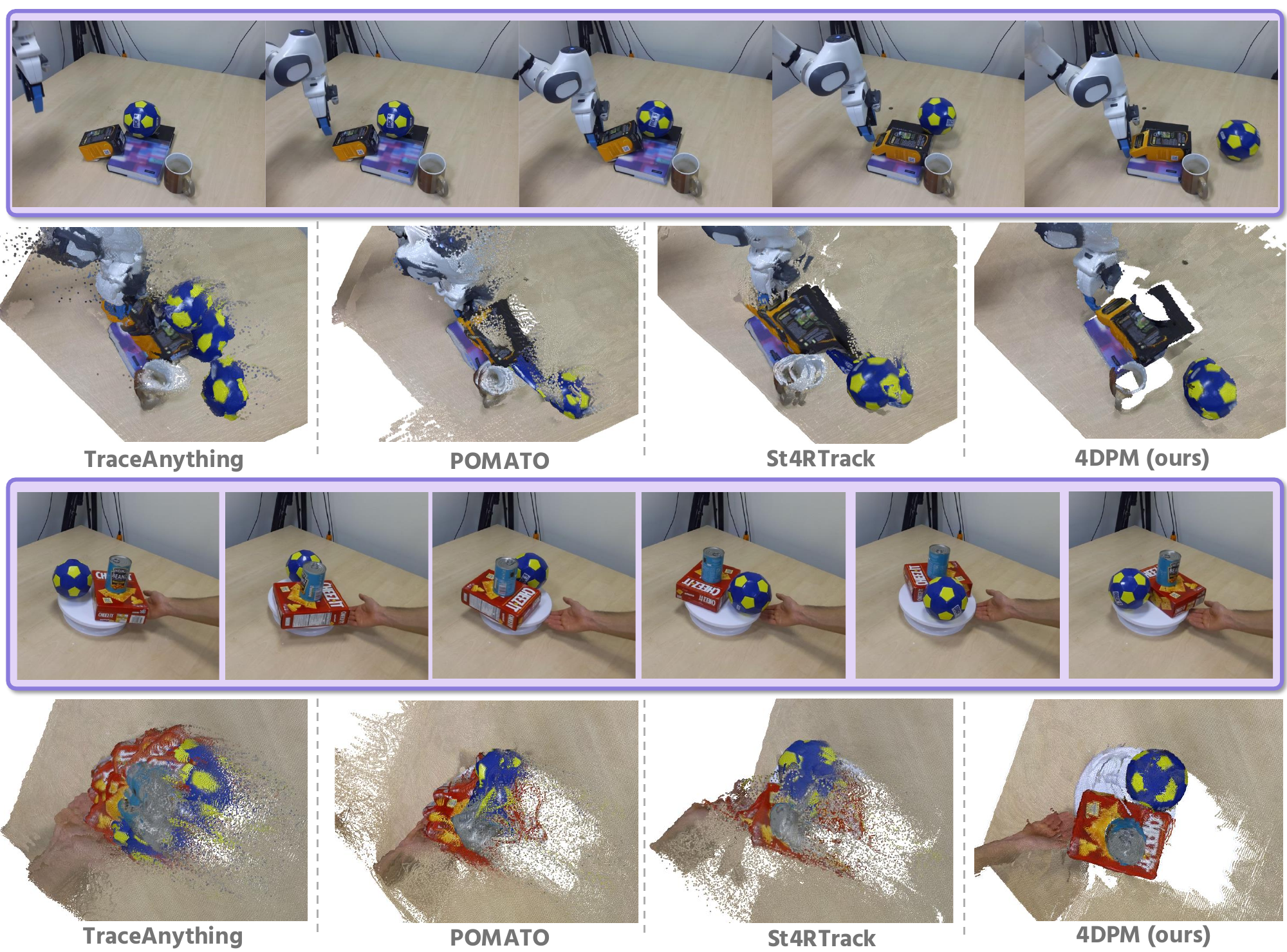}
    \caption{\textbf{Qualitative comparison on Multi-Object dataset.}
Input video frames are shown in purple. Below each video, we visualise all observed point-maps time-warped to the latest timestamp. Our system successfully handles multi-object motion and performs well on particularly challenging objects such as the spinning ball and robot gripper \textbf{(top row)}. We provide a top-down view of multiple objects spinning on a rotating base \textbf{(bottom row)}. Our method correctly aggregates all observations, resulting in complete and accurate object scans.
} 
    \label{fig:comparison}
\end{figure*}

Due to the scarcity of real-world datasets capturing complex multi-object rigid motion, we collected our own dataset using four time-synchronised Azure Kinect cameras. Following our evaluation protocol for HO3D, only one camera stream serves as input to all methods, while all four synchronised views are used to generate more reliable pseudo-ground-truth (pseudo-GT) geometry. Although these cameras include depth sensors, we observed unsatisfactory depth quality for most objects in practice. We therefore employed the metric-aligned feed-forward reconstruction from~\cite{wang2025pi3} on all four synchronised views to obtain pseudo-GT geometry estimates. We provide ground-truth masks for all dynamic scene parts and evaluate using the same geometric metrics as on HO3D, namely accuracy, recall and F-score all evaluated at 1cm threshold.

Quantitative results are presented in \cref{tab:franka}: our method marginally outperforms all other baselines in terms of our main evaluation metric, F-score. Qualitatively, as seen in~\cref{fig:comparison} our method produces significantly more accurate and complete reconstructions than others.

\begin{table*}[h]
\resizebox{\linewidth}{!}{%
    \begin{tabular}{l|ccccccc|c||cc}
    \hline
                              & BallPush & BoxSpin & MultiObj1 & MultiObj2 & PanStir & Spin1  & Spin2  & Mean & Precision & Recall\\ 
    \hline
        $\pi^3$ last view     & 0.3627   & 0.5296  & 0.6470     & \underline{0.8318}    & 0.1898  & 0.4711 & 0.5179 & 0.5071     &  \textbf{0.8837}  & 0.3707 \\ 
        $\pi^3$               & 0.2176   & 0.6128  & \underline{0.6828}    & 0.6900   & 0.1640   & 0.3987 & \underline{0.5934}   & 0.4799   &   0.3637 &  0.7382\\ \hline
        St4track              & 0.3968   & \underline{0.7955}  & 0.4514    & 0.6608    & 0.0401  & 0.4778 & 0.4189 & 0.4630     &    0.3585   &   0.6792  \\ 
        POMATO                & 0.4311   & 0.7175  & 0.6541    & 0.6373    & 0.4597  & \underline{0.6225} & 0.5827 & \underline{0.5864}     &  0.4668  &     \underline{0.8071}  \\
        TraceAny              & \underline{0.4581}   & 0.6876  & 0.3191    & 0.6175    & \underline{0.4616}  & 0.3706 & 0.4578 & 0.4817     &  0.3773  &     0.6946  \\ \hline
        Ours                  & \textbf{0.7683}   & \textbf{0.9179}  & \textbf{0.8226}    & \textbf{0.8903}    & \textbf{0.6359}  & \textbf{0.8544} & \textbf{0.6746} & \textbf{0.7948} &  \underline{0.7195} & \textbf{0.9000} \\ \hline
    \end{tabular}
}
    \label{tab:franka}
    \caption{\textbf{Quantitative evaluation on Multi-Object dataset.} We report F-score (threshold at 1 cm) per sequence for all methods. Average F-score, precision, and recall across all sequences are also reported. Our method marginally outperforms all baselines in terms of F-score multi-object dynamic reconstruction, providing the best balance between completeness and accuracy. Best is highlighted as \textbf{bold}, while second-best is \underline{underscored}.}

\end{table*}

\subsection{Object permanence}
\begin{figure*}[h]
    \centering
    \includegraphics[width=\linewidth]{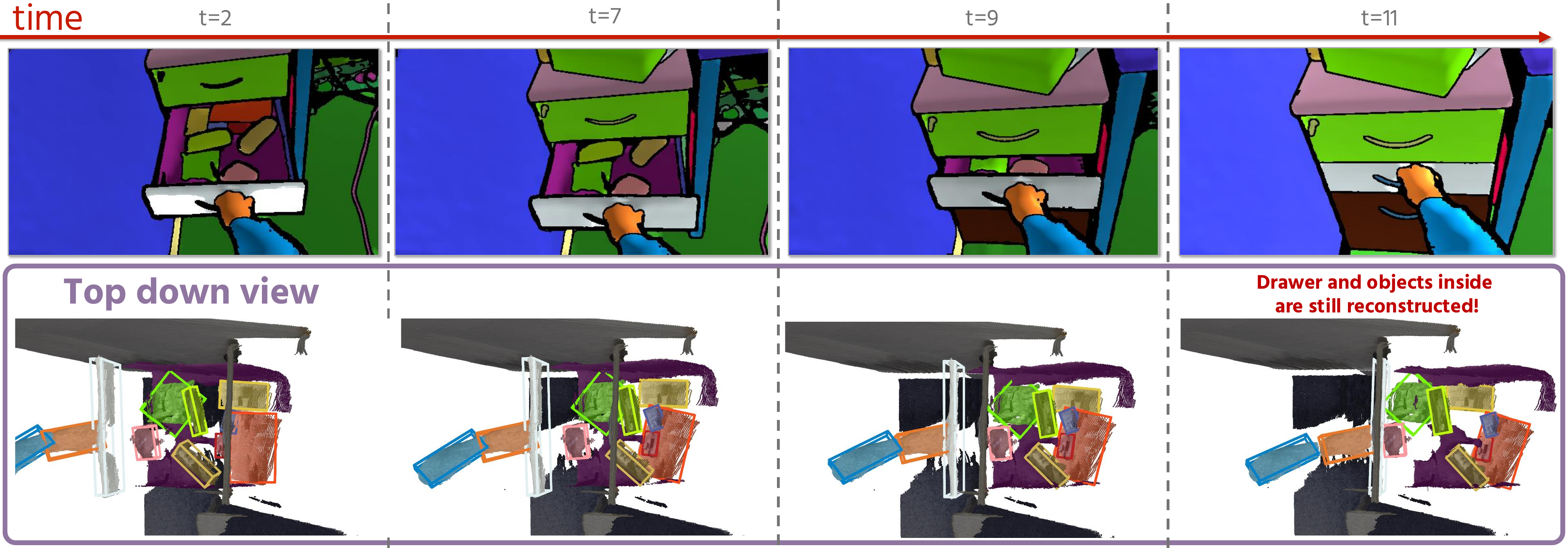}
    \caption{\textbf{Object permanence capabilities.} 
    In \textbf{(top row)} we show input frames of a closing the drawer sequence. The resulting reconstruction estimated with 4DPM from the top-down view in the \textbf{(bottom row)}. When the drawer is fully closed (rightmost column), our method still reconstructs objects inside the drawer and the drawer body, despite it being completely occluded. This showcases object permanence capabilities of 4DPM. 
    The top of the drawer is removed from reconstruction for better viewing.} 
    \vspace{-8pt}
    \label{fig:drawer}
\end{figure*}

Our primitive-based representation naturally allows us to reason about spatial relationships between moving scene parts. This capability proves especially useful when objects become occluded, due to motion.
In Figure~\ref{fig:drawer}, we demonstrate our system operating on a closing drawer sequence. We show the top-down view of our 4D reconstruction. Initially, objects inside the drawer are visible and can be ordinarily reconstructed. As the drawer closes, these objects gradually become occluded, yet their motion and position can be inferred from the drawer front's movement. Our primitive-based 3D representation, coupled with the motion segmentation technique described in~\cref{sec:motionseg}, enables such capability.

Based on motion segmentation, our system groups the drawer body with its front. Importantly, the objects inside are not in direct contact with the drawer front; hence their motion must be transitively inferred through the drawer body. Consequently, these occluded objects remain spatially associated and motion-grouped with the drawer, preserving them in the reconstruction.

To the best of the authors' knowledge, this is the first system to demonstrate such capabilities from casual monocular videos. We strongly encourage reviewers to consult the supplementary materials for video demonstrations of this capability.

\section{Limitations}

Our system assumes that each primitive is rigid and thus cannot represent more intricate non-rigid deformations. 
Extending the method to handle such deformations whilst maintaining computational efficiency remains an important direction for future work. Additionally, incremental mapping capabilities, where the scene representation is built and updated over extended sequences, have yet to be explored.

\section{Conclusion}

We have presented 4D Primitive-Mâché, a novel method for monocular 4D dense scene reconstruction from casual RGB videos. Wielding a primitive-based motion parametrisation, our approach achieves more accurate and complete reconstructions than existing monocular methods on both object scanning and multi-object interaction scenarios. Beyond improved geometric accuracy, our method's persistence mechanism enables spatial memory capabilities, maintaining representations of occluded objects --- a capability not previously demonstrated in monocular reconstruction systems. 

Our primitive-based formulation dramatically reduces the dimensionality of the dynamic reconstruction problem while preserving expressiveness for complex scene motion, opening new possibilities for robust scene reconstruction and understanding from monocular video.

\paragraph{Acknowledgments}
Research presented in this paper was supported by Dyson Technology Ltd. The authors would like to thank Eric Dexheimer and other members of the Dyson Robotics Lab for insightful discussions.

{
    \small
    \bibliographystyle{ieeenat_fullname}
    \bibliography{robotvision}

\begin{thebibliography}{43}
\providecommand{\natexlab}[1]{#1}
\providecommand{\url}[1]{\texttt{#1}}
\expandafter\ifx\csname urlstyle\endcsname\relax
  \providecommand{\doi}[1]{doi: #1}\else
  \providecommand{\doi}{doi: \begingroup \urlstyle{rm}\Url}\fi

\bibitem[Akhter et~al.(2008)Akhter, Sheikh, Khan, and Kanade]{akhter2008nonrigid}
Ijaz Akhter, Yaser Sheikh, Sohaib Khan, and Takeo Kanade.
\newblock Nonrigid structure from motion in trajectory space.
\newblock In \emph{{Neural Information Processing Systems ({NeurIPS})}}, 2008.

\bibitem[Bloesch et~al.(2018)Bloesch, Czarnowski, Clark, Leutenegger, and Davison]{Bloesch:etal:CVPR2018}
M. Bloesch, J. Czarnowski, R. Clark, S. Leutenegger, and A.~J. Davison.
\newblock {CodeSLAM} --- learning a compact, optimisable representation for dense visual {SLAM}.
\newblock In \emph{{Proceedings of the {IEEE} Conference on Computer Vision and Pattern Recognition ({CVPR})}}, 2018.

\bibitem[Bregler et~al.(2000)Bregler, Hertzmann, and Biermann]{Bregler2000RecoveringN3}
Christoph Bregler, Aaron Hertzmann, and Henning Biermann.
\newblock Recovering non-rigid 3d shape from image streams.
\newblock In \emph{{Proceedings of the {IEEE} Conference on Computer Vision and Pattern Recognition ({CVPR})}}, 2000.

\bibitem[Campos et~al.(2021)Campos, Elvira, Gomez, Montiel, and Tardos]{ORBSLAM3_TRO}
Carlos Campos, Richard Elvira, Juan~J. Gomez, Jose M.~M. Montiel, and Juan~D. Tardos.
\newblock {ORB-SLAM3}: An accurate open-source library for visual, visual-inertial and multi-map {SLAM}.
\newblock \emph{IEEE Transactions on Robotics}, 2021.

\bibitem[Costeira and Kanade(1995)]{costeira1995multi}
Joao Costeira and Takeo Kanade.
\newblock A multi-body factorization method for motion analysis.
\newblock In \emph{{Proceedings of the International Conference on Computer Vision ({ICCV})}}, 1995.

\bibitem[Czarnowski et~al.(2020)Czarnowski, Laidlow, Clark, and Davison]{Czarnowski:etal:RAL2020}
J. Czarnowski, T. Laidlow, R. Clark, and A.~J. Davison.
\newblock Deepfactors: Real-time probabilistic dense monocular {SLAM}.
\newblock In \emph{{{IEEE} Robotics and Automation Letters}}, 2020.

\bibitem[Davison et~al.(2007)Davison, Molton, Reid, and Stasse]{Davison:etal:PAMI2007}
A.~J. Davison, N.~D. Molton, I. Reid, and O. Stasse.
\newblock {{MonoSLAM}: Real-Time Single Camera {SLAM}}.
\newblock \emph{{{IEEE} Transactions on Pattern Analysis and Machine Intelligence ({PAMI})}}, 2007.

\bibitem[Dexheimer and Davison(2024)]{dexheimer2024compact}
Eric Dexheimer and Andrew~J Davison.
\newblock Como: Compact mapping and odometry.
\newblock In \emph{European Conference on Computer Vision}, pages 349--365. Springer, 2024.

\bibitem[Feng* et~al.(2025)Feng*, Zhang*, Wang, Ye, Yu, Black, Darrell, and Kanazawa]{st4rtrack2025}
Haiwen Feng*, Junyi Zhang*, Qianqian Wang, Yufei Ye, Pengcheng Yu, Michael~J. Black, Trevor Darrell, and Angjoo Kanazawa.
\newblock St4rtrack: Simultaneous 4d reconstruction and tracking in the world.
\newblock In \emph{{Proceedings of the {IEEE} Conference on Computer Vision and Pattern Recognition ({CVPR})}}, 2025.

\bibitem[Hampali et~al.(2020)Hampali, Rad, Oberweger, and Lepetit]{hampali2020honnotate}
Shreyas Hampali, Mahdi Rad, Markus Oberweger, and Vincent Lepetit.
\newblock Honnotate: A method for 3d annotation of hand and object poses.
\newblock In \emph{CVPR}, 2020.

\bibitem[Harley et~al.(2025)Harley, You, Sun, Zheng, Raghuraman, Gu, Liang, Chu, Dave, You, et~al.]{harley2025alltracker}
Adam~W Harley, Yang You, Xinglong Sun, Yang Zheng, Nikhil Raghuraman, Yunqi Gu, Sheldon Liang, Wen-Hsuan Chu, Achal Dave, Suya You, et~al.
\newblock Alltracker: Efficient dense point tracking at high resolution.
\newblock In \emph{{Proceedings of the International Conference on Computer Vision ({ICCV})}}, 2025.

\bibitem[Huang et~al.(2025)Huang, Zhou, Rabeti, Korovko, Ling, Ren, Shen, Gao, Slepichev, Lin, et~al.]{huang2025vipe}
Jiahui Huang, Qunjie Zhou, Hesam Rabeti, Aleksandr Korovko, Huan Ling, Xuanchi Ren, Tianchang Shen, Jun Gao, Dmitry Slepichev, Chen-Hsuan Lin, et~al.
\newblock Vipe: Video pose engine for 3d geometric perception.
\newblock \emph{arXiv preprint arXiv:2508.10934}, 2025.

\bibitem[Kingma and Ba(2014)]{Kingma2014AdamAM}
Diederik~P. Kingma and Jimmy Ba.
\newblock Adam: A method for stochastic optimization.
\newblock \emph{CoRR}, abs/1412.6980, 2014.

\bibitem[Klein and Murray(2007)]{Klein:Murray:ISMAR2007}
G. Klein and D.~W. Murray.
\newblock {Parallel Tracking and Mapping for Small {AR} Workspaces}.
\newblock In \emph{{Proceedings of the International Symposium on Mixed and Augmented Reality ({ISMAR})}}, 2007.

\bibitem[Knapitsch et~al.(2017)Knapitsch, Park, Zhou, and Koltun]{Knapitsch2017}
Arno Knapitsch, Jaesik Park, Qian-Yi Zhou, and Vladlen Koltun.
\newblock Tanks and temples: Benchmarking large-scale scene reconstruction.
\newblock \emph{ACM Transactions on Graphics}, 36\penalty0 (4), 2017.

\bibitem[Kopf et~al.(2021)Kopf, Rong, and Huang]{kopf2021robust}
Johannes Kopf, Xuejian Rong, and Jia-Bin Huang.
\newblock Robust consistent video depth estimation.
\newblock In \emph{Proceedings of the IEEE/CVF Conference on Computer Vision and Pattern Recognition}, pages 1611--1621, 2021.

\bibitem[Kumar et~al.(2017)Kumar, Dai, and Li]{kumar2017monocular}
Suryansh Kumar, Yuchao Dai, and Hongdong Li.
\newblock Monocular dense 3d reconstruction of a complex dynamic scene from two perspective frames.
\newblock In \emph{Proceedings of the IEEE international conference on computer vision}, pages 4649--4657, 2017.

\bibitem[Kumar et~al.(2019)Kumar, Dai, and Li]{kumar2019superpixel}
Suryansh Kumar, Yuchao Dai, and Hongdong Li.
\newblock Superpixel soup: Monocular dense 3d reconstruction of a complex dynamic scene.
\newblock \emph{IEEE transactions on pattern analysis and machine intelligence}, 43\penalty0 (5):\penalty0 1705--1717, 2019.

\bibitem[Li et~al.(2025)Li, Tucker, Cole, Wang, Jin, Ye, Kanazawa, Holynski, and Snavely]{li2025megasam}
Zhengqi Li, Richard Tucker, Forrester Cole, Qianqian Wang, Linyi Jin, Vickie Ye, Angjoo Kanazawa, Aleksander Holynski, and Noah Snavely.
\newblock Megasam: Accurate, fast and robust structure and motion from casual dynamic videos.
\newblock In \emph{Proceedings of the Computer Vision and Pattern Recognition Conference}, pages 10486--10496, 2025.

\bibitem[Liu et~al.(2025)Liu, Xiao, Chen, Feng, Tai, Tang, and Kang]{liu2025trace}
Xinhang Liu, Yuxi Xiao, Donny~Y Chen, Jiashi Feng, Yu-Wing Tai, Chi-Keung Tang, and Bingyi Kang.
\newblock Trace anything: Representing any video in 4d via trajectory fields.
\newblock \emph{arXiv preprint arXiv:2510.13802}, 2025.

\bibitem[Matsuki et~al.(2021)Matsuki, Scona, Czarnowski, and Davison]{Matsuki:etal:RAL2021}
H. Matsuki, R. Scona, J. Czarnowski, and A.~J. Davison.
\newblock {CodeMapping}: Real-time dense mapping for sparse {SLAM} using compact scene representations.
\newblock \emph{{{IEEE} Robotics and Automation Letters}}, 6\penalty0 (4):\penalty0 7105--7112, 2021.

\bibitem[Mazur et~al.(2024)Mazur, Bae, and Davison]{Mazur:etal:CVPR2024}
Kirill Mazur, Gwangbin Bae, and Andrew Davison.
\newblock {SuperPrimitive}: Scene reconstruction at a primitive level.
\newblock In \emph{IEEE/CVF Conference on Computer Vision and Pattern Recognition (CVPR)}, 2024.

\bibitem[Murai et~al.(2025)Murai, Dexheimer, and Davison]{murai2025mast3r}
Riku Murai, Eric Dexheimer, and Andrew~J Davison.
\newblock Mast3r-slam: Real-time dense slam with 3d reconstruction priors.
\newblock In \emph{{Proceedings of the {IEEE} Conference on Computer Vision and Pattern Recognition ({CVPR})}}, 2025.

\bibitem[Newcombe et~al.(2015)Newcombe, Fox, and Seitz]{Newcombe:etal:CVPR2015}
Richard~A Newcombe, Dieter Fox, and Steven~M Seitz.
\newblock Dynamicfusion: Reconstruction and tracking of non-rigid scenes in real-time.
\newblock In \emph{{Proceedings of the {IEEE} Conference on Computer Vision and Pattern Recognition ({CVPR})}}, 2015.

\bibitem[Ravi et~al.(2024)Ravi, Gabeur, Hu, Hu, Ryali, Ma, Khedr, R{\"a}dle, Rolland, Gustafson, Mintun, Pan, Alwala, Carion, Wu, Girshick, Doll{\'a}r, and Feichtenhofer]{ravi2024sam2}
Nikhila Ravi, Valentin Gabeur, Yuan-Ting Hu, Ronghang Hu, Chaitanya Ryali, Tengyu Ma, Haitham Khedr, Roman R{\"a}dle, Chloe Rolland, Laura Gustafson, Eric Mintun, Junting Pan, Kalyan~Vasudev Alwala, Nicolas Carion, Chao-Yuan Wu, Ross Girshick, Piotr Doll{\'a}r, and Christoph Feichtenhofer.
\newblock Sam 2: Segment anything in images and videos.
\newblock \emph{arXiv preprint arXiv:2408.00714}, 2024.

\bibitem[R{\"u}nz and Agapito(2017)]{Runz::Agapito::ICRA2017}
Martin R{\"u}nz and Lourdes Agapito.
\newblock Co-fusion: Real-time segmentation, tracking and fusion of multiple objects.
\newblock In \emph{{Proceedings of the {IEEE} International Conference on Robotics and Automation ({ICRA})}}, 2017.

\bibitem[Salas-Moreno et~al.(2013)Salas-Moreno, Newcombe, Strasdat, Kelly, and Davison]{Salas-Moreno:etal:CVPR2013}
R.~F. Salas-Moreno, R.~A. Newcombe, H. Strasdat, P.~H.~J. Kelly, and A.~J. Davison.
\newblock {{SLAM++}: Simultaneous Localisation and Mapping at the Level of Objects}.
\newblock In \emph{{Proceedings of the {IEEE} Conference on Computer Vision and Pattern Recognition ({CVPR})}}, 2013.

\bibitem[Sch\"{o}nberger and Frahm(2016)]{schoenberger:etal:CVPR16}
Johannes~Lutz Sch\"{o}nberger and Jan-Michael Frahm.
\newblock Structure-from-motion revisited.
\newblock In \emph{{Proceedings of the {IEEE} Conference on Computer Vision and Pattern Recognition ({CVPR})}}, 2016.

\bibitem[Sola et~al.(2018)Sola, Deray, and Atchuthan]{sola2018micro}
Joan Sola, Jeremie Deray, and Dinesh Atchuthan.
\newblock A micro lie theory for state estimation in robotics.
\newblock \emph{arXiv preprint arXiv:1812.01537}, 2018.

\bibitem[Strasdat et~al.(2010)Strasdat, Montiel, and Davison]{Strasdat:etal:RSS2010}
H. Strasdat, J.~M.~M. Montiel, and A.~J. Davison.
\newblock {Scale Drift-Aware Large Scale Monocular {SLAM}}.
\newblock In \emph{{Proceedings of Robotics: Science and Systems ({RSS})}}, 2010.

\bibitem[Sucar et~al.(2025)Sucar, Lai, Insafutdinov, and Vedaldi]{sucar2025dynamic}
Edgar Sucar, Zihang Lai, Eldar Insafutdinov, and Andrea Vedaldi.
\newblock Dynamic point maps: A versatile representation for dynamic 3d reconstruction.
\newblock In \emph{{Proceedings of the International Conference on Computer Vision ({ICCV})}}, 2025.

\bibitem[Tresadern and Reid(2005)]{tresadern2005articulated}
Phil Tresadern and Ian Reid.
\newblock Articulated structure from motion by factorization.
\newblock In \emph{{Proceedings of the {IEEE} Conference on Computer Vision and Pattern Recognition ({CVPR})}}, 2005.

\bibitem[Triggs et~al.(1999)Triggs, McLauchlan, Hartley, and Fitzgibbon]{Triggs:etal:VISALG1999}
B. Triggs, P. McLauchlan, R. Hartley, and A. Fitzgibbon.
\newblock {Bundle Adjustment --- A Modern Synthesis}.
\newblock In \emph{{Proceedings of the International Workshop on Vision Algorithms, in association with {ICCV}}}, 1999.

\bibitem[Umeyama(1991)]{Umeyama1991LeastSquaresEO}
Shinji Umeyama.
\newblock Least-squares estimation of transformation parameters between two point patterns.
\newblock \emph{IEEE Trans. Pattern Anal. Mach. Intell.}, 13:\penalty0 376--380, 1991.

\bibitem[Wang et~al.(2025{\natexlab{a}})Wang, Chen, Karaev, Vedaldi, Rupprecht, and Novotny]{wang2025vggt}
Jianyuan Wang, Minghao Chen, Nikita Karaev, Andrea Vedaldi, Christian Rupprecht, and David Novotny.
\newblock Vggt: Visual geometry grounded transformer.
\newblock In \emph{{Proceedings of the {IEEE} Conference on Computer Vision and Pattern Recognition ({CVPR})}}, 2025{\natexlab{a}}.

\bibitem[Wang et~al.(2024)Wang, Leroy, Cabon, Chidlovskii, and Revaud]{wang2024dust3r}
Shuzhe Wang, Vincent Leroy, Yohann Cabon, Boris Chidlovskii, and Jerome Revaud.
\newblock Dust3r: Geometric 3d vision made easy.
\newblock In \emph{Proceedings of the IEEE/CVF Conference on Computer Vision and Pattern Recognition}, pages 20697--20709, 2024.

\bibitem[Wang et~al.(2025{\natexlab{b}})Wang, Zhou, Zhu, Chang, Zhou, Li, Chen, Pang, Shen, and He]{wang2025pi3}
Yifan Wang, Jianjun Zhou, Haoyi Zhu, Wenzheng Chang, Yang Zhou, Zizun Li, Junyi Chen, Jiangmiao Pang, Chunhua Shen, and Tong He.
\newblock {$\pi^3$}: Scalable permutation-equivariant visual geometry learning, 2025{\natexlab{b}}.

\bibitem[Xu et~al.(2019)Xu, Li, Tzoumanikas, Bloesch, Davison, and Leutenegger]{Xu:etal:ICRA2019}
Binbin Xu, Wenbin Li, Dimos Tzoumanikas, Michael Bloesch, Andrew Davison, and Stefan Leutenegger.
\newblock {MID-Fusion}: Octree-based object-level multi-instance dynamic slam.
\newblock In \emph{{Proceedings of the {IEEE} International Conference on Robotics and Automation ({ICRA})}}, 2019.

\bibitem[Zhang et~al.(2025{\natexlab{a}})Zhang, Herrmann, Hur, Jampani, Darrell, Cole, Sun, and Yang]{zhang2024monst3r}
Junyi Zhang, Charles Herrmann, Junhwa Hur, Varun Jampani, Trevor Darrell, Forrester Cole, Deqing Sun, and Ming-Hsuan Yang.
\newblock Monst3r: A simple approach for estimating geometry in the presence of motion.
\newblock In \emph{{Proceedings of the International Conference on Learning Representations ({ICLR})}}, 2025{\natexlab{a}}.

\bibitem[Zhang et~al.(2025{\natexlab{b}})Zhang, Ge, Tian, Xu, Chen, Lv, and Shen]{Zhang_2025_ICCV}
Songyan Zhang, Yongtao Ge, Jinyuan Tian, Guangkai Xu, Hao Chen, Chen Lv, and Chunhua Shen.
\newblock Pomato: Marrying pointmap matching with temporal motions for dynamic 3d reconstruction.
\newblock In \emph{{Proceedings of the International Conference on Computer Vision ({ICCV})}}, 2025{\natexlab{b}}.

\bibitem[Zhang et~al.(2022)Zhang, Cole, Li, Rubinstein, Snavely, and Freeman]{zhang2022structure}
Zhoutong Zhang, Forrester Cole, Zhengqi Li, Michael Rubinstein, Noah Snavely, and William~T Freeman.
\newblock Structure and motion from casual videos.
\newblock In \emph{European Conference on Computer Vision}, pages 20--37. Springer, 2022.

\bibitem[Zheng et~al.(2025)Zheng, Zhang, Salehi, Gao, Iyengar, Kobori, Kong, and Krishna]{zheng2025one}
Chenhao Zheng, Jieyu Zhang, Mohammadreza Salehi, Ziqi Gao, Vishnu Iyengar, Norimasa Kobori, Quan Kong, and Ranjay Krishna.
\newblock One trajectory, one token: Grounded video tokenization via panoptic sub-object trajectory.
\newblock In \emph{{Proceedings of the International Conference on Computer Vision ({ICCV})}}, 2025.

\bibitem[Zhou et~al.(2025)Zhou, Wang, Chen, Chang, Beaudouin, Zhan, Liang, and Wang]{zhou2025page}
Kaichen Zhou, Yuhan Wang, Grace Chen, Xinhai Chang, Gaspard Beaudouin, Fangneng Zhan, Paul~Pu Liang, and Mengyu Wang.
\newblock Page-4d: Disentangled pose and geometry estimation for 4d perception.
\newblock \emph{arXiv preprint arXiv:2510.17568}, 2025.

\end{thebibliography}
}

\newpage

\section{Analytical Jacobian Derivation}
We provide analytical Jacobians for a single pairwise residual of an object. In practice, the Hessian is block-diagonal with respect to objects. 

Given relative residuals $r = T_j^{-1} T_i X_i - \reallywidehat{X_j}$, we derive its analytical right Jacobians $\mathbf{J} = [ \mathbf{J} _{T_i} | \mathbf{J}_{T_j}]$ with respect to object poses $T_i$ and $T_j$.
    

Let $Z = T_j^{-1} T_i$, then:
\begin{equation}
    \frac{dZ}{dT_i} = \operatorname{Id}
\end{equation}

\begin{equation}
\frac{dZ}{dT_j} = - \operatorname{Ad} \left(Z^{-1} \right)
\end{equation}

From~\cite{sola2018micro}, if $Z =\begin{bmatrix} 
R_Z & t_Z \\
0 & 1  
\end{bmatrix}$, $R_Z \in \SO, t_Z \in \RR^3$

and $\operatorname{Act}$ is the action operator of $\SE$ group on $\RR^3$, i.e $\operatorname{Act}(Z, X) = R_Z X + t_Z$, then:

\begin{equation}
\frac{d \operatorname{Act}(Z, X)}{dZ} = \begin{bmatrix}
     R_Z  &  -R_Z [X]_{\times}
\end{bmatrix}
\end{equation}

Then, using chainrule,

\begin{equation}
\begin{split}
&\frac{dr}{dT_i} = \frac{d(Z X_i - \reallywidehat{X_j})}{dT_i} = \\
& = \frac{d \operatorname{Act}(Z, X)}{dZ} \frac{dZ}{d T_i} =  \begin{bmatrix}
     R_Z  &  -R_Z [X_i]_{\times}
     \end{bmatrix}
\end{split}
\end{equation}

\begin{equation}
\begin{split}
&\frac{dr}{dT_j} = \frac{d(Z X_i - \reallywidehat{X_j})}{dT_j} = \frac{d \operatorname{Act}(Z, X_i)}{dZ} \frac{dZ}{d T_j} = \\ 
& = \begin{bmatrix}
     R_Z  &  -R_Z [X_i]_{\times}
     \end{bmatrix}
\left(- \operatorname{Ad} \left(Z^{-1} \right) \right)
\end{split}
\end{equation}

The last equation can be simplified further. Recall that $Z^{-1} = \begin{bmatrix}
    R_Z^T  & - R_Z^T t_Z \\
    0  & 1 
\end{bmatrix}$ and, therefore, its adjoint is: 
\begin{equation}
    \operatorname{Ad}(Z^{-1}) =  \begin{bmatrix}
    R_Z^T & [-R_Z t_Z]_{\times} R_Z^T \\
    0 & R_Z^T
\end{bmatrix}
\end{equation}

Then the second-block column of $\frac{dr}{dT_j}$ corresponding to rotation is: 
\begin{equation}
\begin{split}    
   & - R_Z [-R_Z t_Z]_{\times} R_Z^T + R_Z [X_i]_{\times} R_Z^T = \\
   & = [t_Z]_{\times} + [R_Z X_i]_{\times} = \\
   & = [R_Z X_i + t_Z ]_{\times} = [\operatorname{Act}(Z, X_i)]_{\times}
\end{split}
\end{equation}

\section{First Order vs Second Order Study}

We compare the performance of our system using Gauss-Newton optimisation against Adam. In~\cref{tab:ablation}, we report the resulting F-score on our Multi-Object dataset for our system compared to a variant using the first-order Adam optimiser~\cite{Kingma2014AdamAM}. Interestingly, even with a generous time budget, the first-order optimiser never converged to the same level of quality, both quantitatively and qualitatively. In our experiments, the translational component of the transformation was often estimated correctly, while the rotational component remained problematic.

\begin{table}[h]
\centering
\begin{tabular}{l|cc}
               &  F-Score   & Time Spent \\ 
\hline
Adam 500 steps &  0.6342     & 20s        \\
Adam 1k steps  &  0.6474    & 40s       \\
Adam 10k steps & 0.7228     &  400s      \\
\hline
Ours (10 steps)        &   0.7843    & 2s  \\
Ours (50 steps)   &  \textbf{0.7948}     &  \textbf{10s}       \\
\hline
\end{tabular}
\label{tab:ablation}
\caption{\textbf{Reconstruction quality of our method with different optimisers and computational budgets.}}
\end{table}

\section{Run-Time Analysis}
\begin{table}[h]
\centering
\begin{tabular}{l|cc}
Component              &  Time Spent \\ 
\hline
Frontend (correspondence)  &     2.3s  \\ 
Frontend ($\pi^3$)         &     5.6s    \\
Frontend (segmentation)    &     42s    \\
\hline 
Frontend (combined)        &     50s    \\
Backend                    &     9.6s     \\
Motion Segmentation        &     2s       \\
\hline
\end{tabular}
\label{tab:performance}
\caption{\textbf{Reconstruction quality of our method with different optimisers and computational budgets.}}
\end{table}

In \cref{tab:performance}, we report a performance breakdown of our system. The frontend dominates overall runtime, with video segmentation being the most expensive component as we propagate masks for all potential objects in the scene (typically around 50-100 objects). Backend performance scales with image resolution: our dense alignment at $512 \times 512$ takes approximately 10 seconds, reducing to 2.5 seconds at $256 \times 256$. 

Further optimisation of the backend is possible, though the frontend now represents the primary opportunity for improvement. Additionally, our current implementation does not include early termination of the optimisation; incorporating this could yield further performance gains.

\end{document}